\documentclass{style/primelab}

\usepackage[utf8]{inputenc} 
\usepackage{hyperref}       
\usepackage{url}            
\usepackage{booktabs}       
\usepackage{amsfonts}       
\usepackage{nicefrac}       
\usepackage{microtype}      
\usepackage{xcolor}         
\usepackage{float}
\usepackage{multirow}

\usepackage{subcaption}
\usepackage{graphicx} 

\usepackage[export]{adjustbox} 

\usepackage{amsmath}

\newlength{\subplotH}
\title{Memorize, Adapt, Ignore: Diagnosing Robot Learning Mechanisms under Training Data Variation}

\author[1 2]{Ke Zhang}
\author[2 3]{Danica J. Sutherland}
\author[1]{Chao Liu}

\affiliation[1]{PRIME Robotics Lab, Department of Mechanical Engineering, The University of British Columbia}
\affiliation[2]{Department of Computer Science, University of British Columbia}
\affiliation[3]{Alberta Machine Intelligence Institute}

\abstract{
Training data variation, whether through designing a domain randomization (DR) scheme in simulation or curating demonstrations for imitation learning, is a primary lever for improving the robustness of robotic manipulation policies.
Yet its underlying mechanisms remain poorly understood, and practitioners typically select randomization parameters through expensive trial and error. We investigate these mechanisms through a series of case studies,
randomizing object size, color, and type as well as scene lighting and linguistic prompts
across settings including
pick-and-place RL in ManiSkill
and fine-tuning of vision-language-action (VLA) models on LIBERO and RoboTwin. 
We examine both model behavior and internal representations, using the representation kernel as our primary diagnostic tool. We show that the representation kernel distinguishes a shift in the internal learning mechanism from \textit{memorizing} different situations with insufficient variation (e.g.\ learning what to do for a large cube, and what to do for a small cube) to \textit{adapting} to the situation at hand with sufficient variation.
An kernel-based signal-to-noise ratio also helps distinguish when policies have learned to \emph{ignore} task-irrelevant factors (e.g.\ treating blue and red cubes identically, instead of learning a blue sub-policy and a red sub-policy).
We use these diagnostics to develop practical guidance for designing DR schemes, selecting models, and detecting shortcut learning. We further compare different kinds of representations and validate our findings with real-world hardware experiments using ACT-based imitation learning.
}


\begin{document}
\maketitle

\section{Introduction}
\label{sec:intro}
Data is scarce and expensive in robotics. Practitioners therefore augment training data with variation, for example through domain randomization in simulation \citep{tobin2017domainrandomizationtransferringdeep,peng2018sim,Tremblay_2018_CVPR_Workshops} or diverse demonstrations and observation augmentation in imitation learning \citep{robomimic2021, lin2025data}. Even after injecting data variation, training data rarely cover the full space of conditions encountered at evaluation. Therefore, a central question is not only how well a policy performs on in-distribution data, but how well it generalizes to out-of-distribution (OOD) data.

In this paper, we argue that diagnosing internal learning mechanisms is particularly important for robot learning because they can strongly affect generalization, while generalization is itself a central challenge in robotics. Furthermore, we argue that injecting training-data variation is not only about covering a broader range of in-distribution data. More importantly, increasing variation can shift a model from memorizing specific cases toward learning shared patterns across task-relevant variation, and from encoding nuisance variation toward learning to ignore it. Diagnosing internal learning mechanisms under different levels and compositions of data variation can therefore help data-design interventions go beyond purely trial-and-error approaches. Such learning behaviors have been studied in related forms in computer vision 
and supervised learning, including overfitting to training samples~\citep{Zhang2017, 
Arpit2017ACL}, representation generalization across conditions~\citep{hermann2020shapes}, 
and invariance to task-irrelevant factors or its failure via shortcut 
learning~\citep{achille2018emergence, Geirhos2020}, yet they have received 
little attention in robot learning.

Besides training data scarcity, evaluation is also challenging in robot learning: it is not only difficult to define appropriate generalization criteria, but rollouts to explore behaviour can be expensive and sometimes infeasible. To address the first problem, inspired by recent research that decomposes variation into interpretable factors~\citep{pumacay2024the,gao2024,xie2024decomposing}, we view generalization as model performance on unseen data along factor axes of interest. To address the second problem, rather than relying only on task-level metrics such as success rate, which treat the model as an end-to-end black box, we examine its internal representations to obtain finer-grained information without requiring complete rollouts. Specifically, we construct data probes sweeping through each factor and feed them through a model to obtain their representations. Rather than examining individual representations in isolation, we compare representations pairwise within and across factor axes. We therefore compute kernels on internal representations, which reflect how similarly a model views different data samples.

With controlled factor probes and representation-kernel metrics, we can diagnose learning mechanisms in an intuitive and efficient way. From the kernel representation view, ``memorize'' means that representations of different values of the same factor remain separated or form distinct clusters, suggesting that the model treats them as separate cases rather than learning a shared pattern. ``Adapt'' means that representations across different values of a task-relevant factor exhibit a shared structure, suggesting that the model has learned a common rule across those conditions. ``Ignore'' means that representations remain relatively insensitive to changes in a task-irrelevant factor, indicating that the model has learned to suppress distinctions along that factor. Plus, we quantify relative sensitivity using Factor Sensitivity Ratio (FSR) and its task-relevant-reference specialization, Signal-to-Noise Ratio (SNR).

Across RL, imitation learning, VLA policies, data-composition experiments, and hardware, we validate the framework in diverse robot-learning settings. For representation kernels, we use the empirical Neural Tangent Kernel (eNTK) by default, which directly describes how training on one sample propagates to others, and compare it with last-hidden-layer and output kernels. For downstream applications, we apply the resulting diagnostics to OOD robustness, shortcut detection~\citep{xing2025shortcut}, and data-design decisions. Together, these results show how representation-level analysis can complement task-level evaluation.

\section{Background and Related Work}
\label{sec:background}
\subsection{Representation Kernels} \label{sec:kernels}
A representation kernel $K(\mathbf{x}, \mathbf{x}') = \langle \phi(\mathbf{x}), \phi(\mathbf{x}') \rangle$
summarizes how a network represents inputs through a feature map $\phi$; different choices
of $\phi$ expose different parts of the network.

The most direct choice is a layer's
activations, e.g., using $\phi$ as the last hidden layer $h(\mathbf{x})$ or the output $f(\mathbf{x})$,
which reveal what features a network encodes or suppresses~\citep{hermann2020shapes}. In robot learning, hidden states of trained policies
encode symbolic task states~\citep{Lu2025ProbingAV}, support behavior
steering~\citep{haon2025mechanistic}, and enable failure
detection~\citep{xu2025can, gu2026safe}. Because robot actions are continuous and
multi-dimensional, the output space is also informative~\citep{agia2024unpacking}. Related
work links properties of pretrained representations to policy
robustness~\citep{Burns2023WhatMakesPVR, jiang2025robots}. Unlike these prior papers, we use
representations to diagnose how a policy handles variation in its training data.

The Neural Tangent Kernel~\citep{jacot2018neural} 
instead uses parameter gradients as features:
\begin{equation}
    K_t(\mathbf{x}, \mathbf{x}') =
    \langle \nabla_\theta f(\theta_t, \mathbf{x}), \nabla_\theta f(\theta_t, \mathbf{x}') \rangle,
\end{equation}
which governs how a gradient step on $\mathbf{x}$ changes the prediction at
$\mathbf{x}'$~\citep[see e.g.][]{Ren_2025}. 
The kernel stays fixed in the lazy-training regime~\citep{chizat2019lazy},
but even in the feature-learning regime of practical networks, the empirical NTK (eNTK)
reflects the current representation and how it
evolves~\citep{he-su:elasticity,fort2020deeplearningversuskernel,wei2022toyrandommatrixmodels,mohamadi2024groktheoreticalanalysisgrokking,Ren_2025}.

Because evaluating the eNTK over all model parameters is expensive, we often restrict the
Jacobian to low-rank adaptation (LoRA) modules~\citep{Hu-lora-iclr-2022}
inserted into selected weight matrices. For a linear layer $z=Wh$ with LoRA
update $W+sBA$, the standard initialization with random $A_0$ and $B_0=0$
gives tangent features proportional to $\delta(\mathbf{x})\otimes
(A_0h(\mathbf{x}))$ for each output coordinate, where $h$ is the layer input
and $\delta$ is the backpropagated sensitivity. This maps the input factor of
the full-weight feature $\delta(\mathbf{x})\otimes h(\mathbf{x})$ into a
low-dimensional space, yielding a cheaper sketch of the selected layers'
kernel contributions. The restricted eNTK is also a diagnostic of the existing
representation: because LoRA often adapts effectively from that
representation, its tangent-feature similarities indicate how an update on
one data point would affect another, even when no LoRA fine-tuning is actually performed.

\subsection{Data Variation and Generalization in Robot Learning}

A central goal in robot learning is generalization to conditions unseen in training, and
the main lever for achieving it is variation in the training data. 
Domain Randomization (DR) serves as a primary paradigm for bridging
the sim-to-real gap by exposing models to a diverse distribution of
simulated environments
\citep{tobin2017domainrandomizationtransferringdeep, peng2018sim,
  Tremblay_2018_CVPR_Workshops}. This approach has been extensively
validated across various tasks, including robotic grasping
\citep{tobin2018domain}, object detection \citep{9916581,
  ren2019domain}, dexterous in-hand manipulation~\citep{Akkaya-RL-solve-rubik-2019,Handa-sim2real-in-hand-manipulation-icra-2023}, and bimanual manipulation \citep{chen2025robotwin20scalabledata}. To move beyond uniform sampling, researchers have introduced structured and context-aware randomization \citep{8794443, Zakharov_2019_ICCV, Yue_2019_ICCV} and investigated the empirical limits of visual sim-to-real transfer \citep{alghonaim2021benchmarking, chen2022understandingdomainrandomizationsimtoreal, Kim_2023_ICCV}. 

In imitation learning, variation comes from demonstration diversity, and a growing body of
work studies how it shapes generalization. Data quality and composition strongly affect
performance~\citep{robomimic2021, belkhale2023data, hejna2024remix}. Generalization differs
across factors of variation~\citep{xie2024decomposing, pumacay2024the} and can compose across
them~\citep{gao2024}. At scale, environment and object diversity drive generalization more
than demonstration count~\citep{lin2025data}, yet not every kind of diversity
helps~\citep{saxena2025matters, shi2025diversity}. 

These studies characterize generalization through outcomes such as success rates. They
show \emph{whether} a policy generalizes across a factor, but not \emph{how}: whether it
memorizes, adapts to, or ignores that factor. We treat DR and demonstration diversity as
instances of \emph{training data variation} and diagnose these mechanisms using metrics on
representation kernels.

\section{Methodology}
\label{sec:setup}

Our pipeline consists of three stages designed to isolate the effects of specific training data variation factors on a model's internal representations.

\textbf{Stage 1: Controlled Training.} We systematically vary one or two target factors while holding the distributions of all other factors fixed, training models across different DR configurations. Consider a robotic pick-and-place task as an example. Some factors are task-relevant (object shape and size, linguistic instructions),
while others are task-irrelevant (object color or background lighting).
When randomizing two factors simultaneously, we structure their joint distribution as either \textit{confounded}, where the factors co-vary (for instance, a specific object color always pairs with a specific size), or \textit{independent}, where all possible combinations of factors occur (a \emph{factorial design}).

\textbf{Stage 2: Probe Construction.} To evaluate the models, we construct specialized probe sets that sweep through a range of values for a target factor while keeping all other input variables static.
The first source of these probes is expert trajectories,
obtained by taking standard expert demonstrations
but re-rendering them with varying factors:
for instance, changing the color of a cube while keeping all positions and actions identical.
The second source is validiation frames,
where we re-render
static frames from the validation set with altered lighting or novel object and background appearances.
By holding all other content strictly constant across these probes, any observed changes in the model can be directly attributed to the specific factor under study.

\textbf{Stage 3: Representation Kernel Diagnostics.} For each considered model checkpoint and probe set, we compute a representation kernel, as discussed in \cref{sec:kernels},
on a subset of inputs.
By default, we use the eNTK:
it aggregates information across the entire network
and characterizes how training on one input propagates to others.
We will empirically compare to hidden-layer and output-layer kernels in Section~\ref{sec:representation-comparison}.
High values
indicate that the model treats $x$ and $x'$ similarly during optimization;
for instance, a gradient step changing the action on $x$ will substantially affect the action on $x'$.
We extract several statistics from this kernel
to measure various effects in data variation.

\begin{figure}[t]
    \centering
    \begin{subfigure}[b]{0.407\textwidth}
        \centering
        \includegraphics[width=\linewidth, height=\subplotH, keepaspectratio]{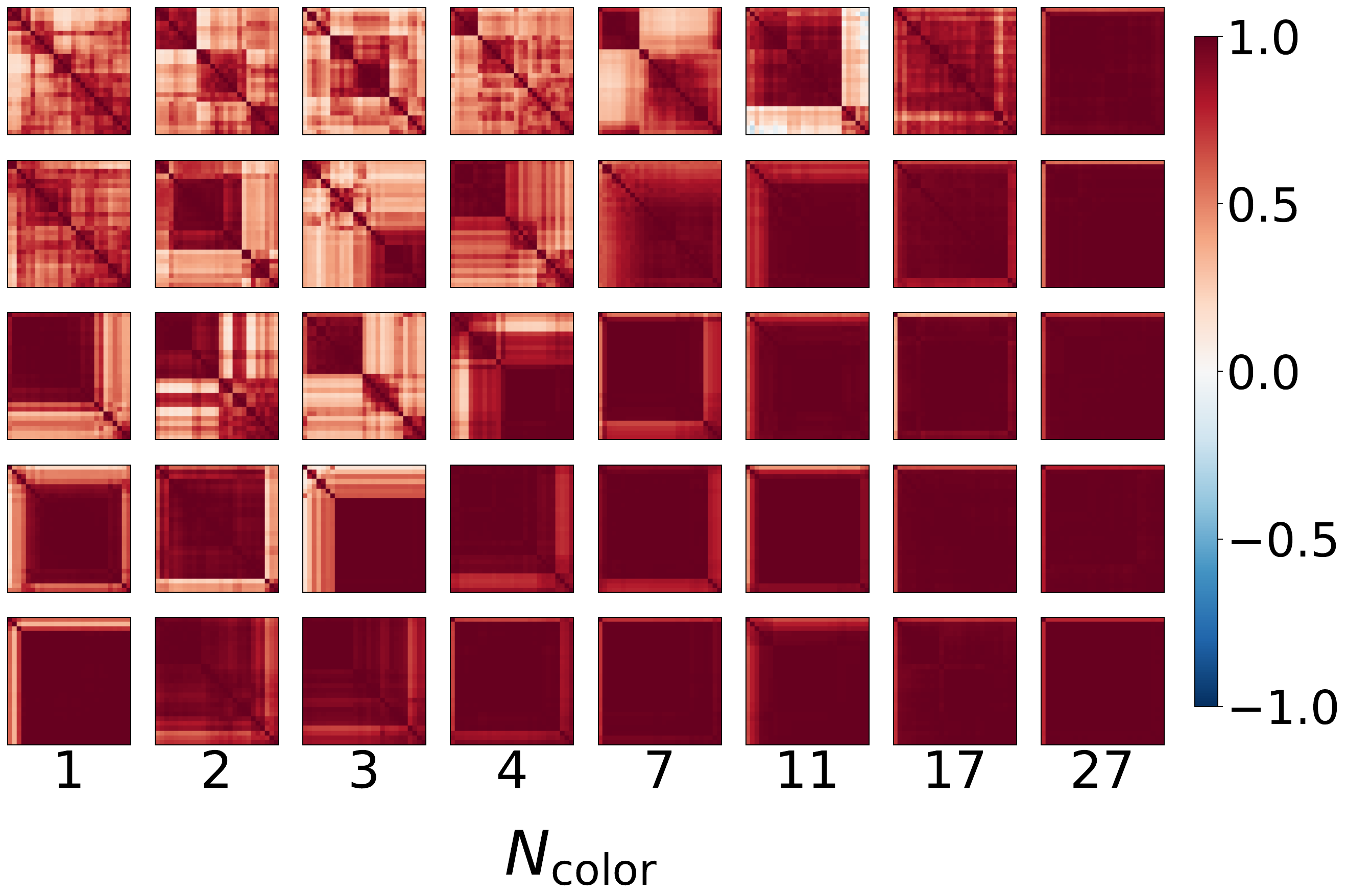}
        \caption{Phase transition heatmap.}
        \label{fig:phase-transition}
    \end{subfigure}
    \hfill
    \begin{subfigure}[b]{0.275\textwidth}
        \centering
        \includegraphics[width=\linewidth, height=\subplotH, keepaspectratio]{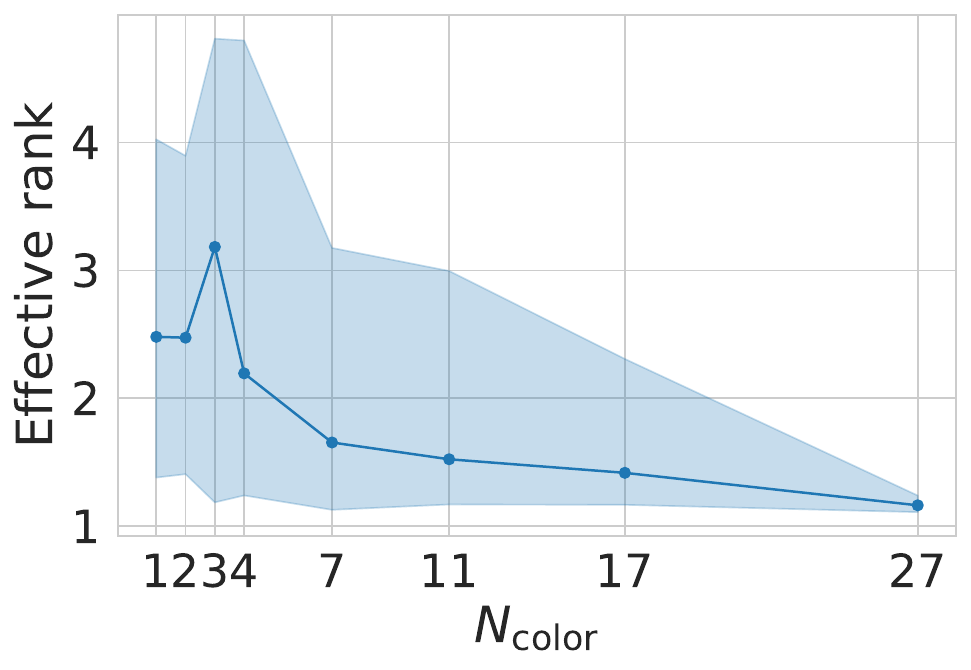}
        \caption{Effective rank.}
        \label{fig:erank}
    \end{subfigure}
    \hfill
    \begin{subfigure}[b]{0.275\textwidth}
        \centering
        \includegraphics[width=\linewidth, height=\subplotH, keepaspectratio]{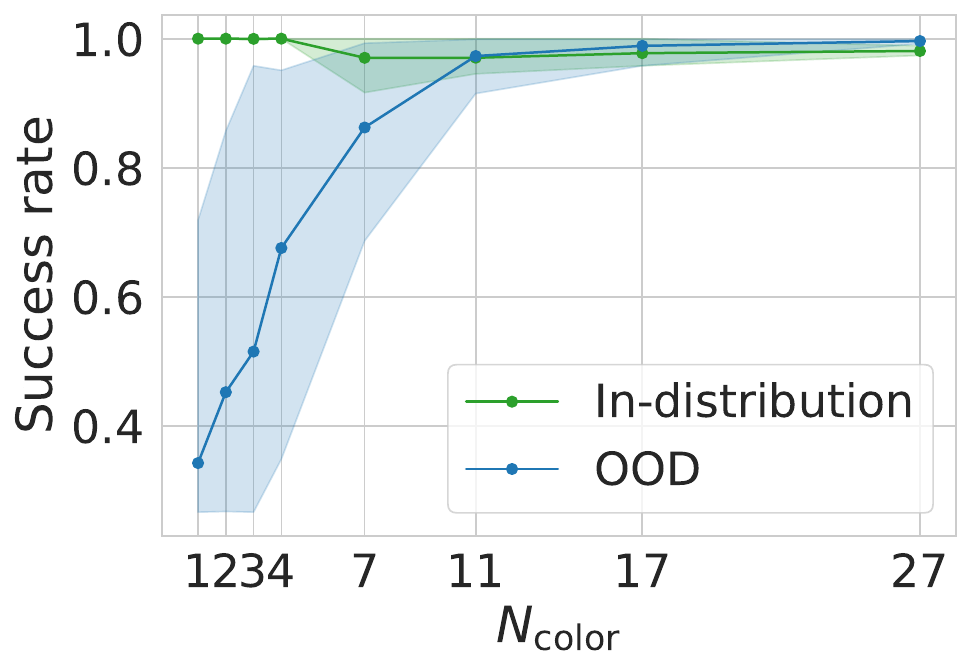}
        \caption{Success rate.}
        \label{fig:success-rate}
    \end{subfigure}

    \vspace{1.5em} 
    

    \hfill
    \begin{subfigure}[b]{0.407\textwidth}
        \centering
        \includegraphics[width=\linewidth, height=\subplotH, keepaspectratio]{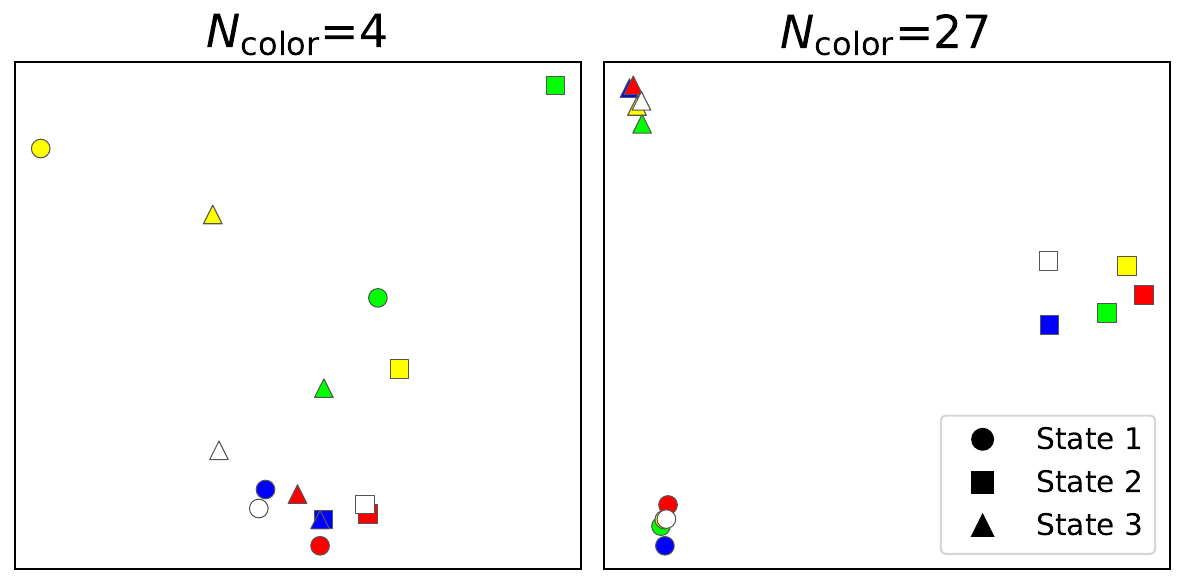}
        \caption{PCA adapt vs ignore.}
        \label{fig:pca-snr}
    \end{subfigure}
    \hfill
    \begin{subfigure}[b]{0.271\textwidth}
        \centering
        \includegraphics[width=\linewidth, height=\subplotH, keepaspectratio]{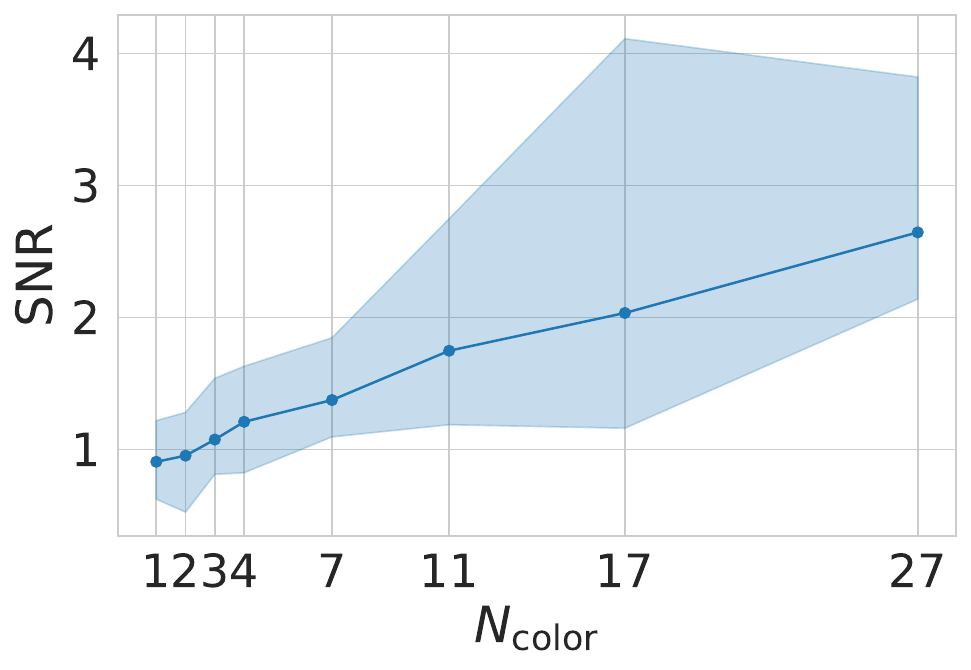}
        \caption{Signal-to-noise ratio.}
        \label{fig:snr-across-n}
    \end{subfigure}
    \begin{subfigure}[b]{0.271\textwidth}
        \centering
        \includegraphics[width=\linewidth, height=\subplotH, keepaspectratio]{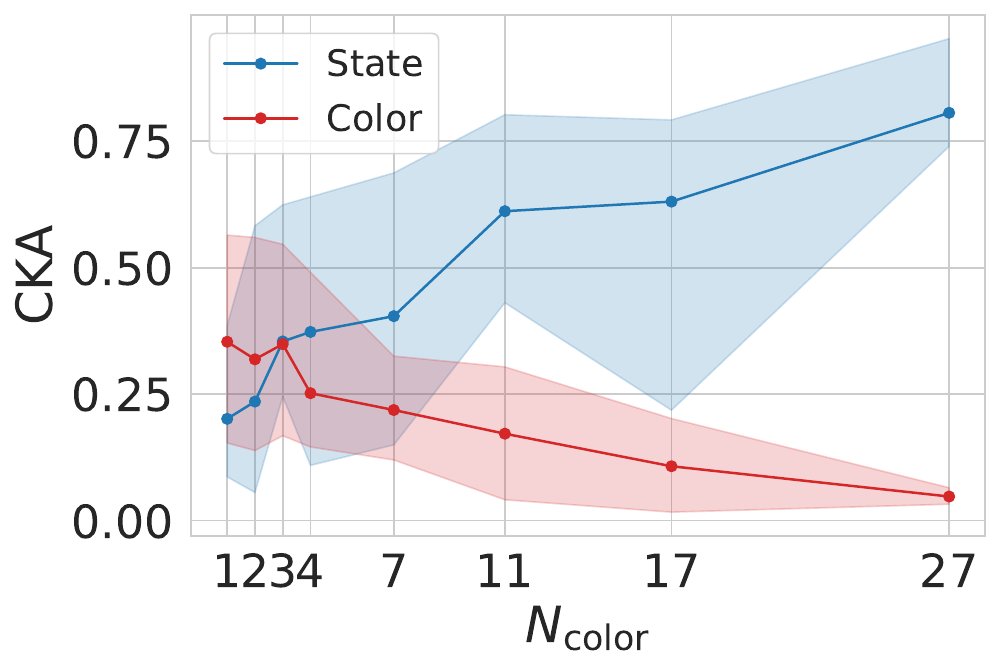}
        \caption{CKA}
        \label{fig:cka-across-n}
    \end{subfigure}
    
    \caption{\textbf{Diagnostic metrics tracking representation kernel phase transitions.} As the number of randomized colors increases, the model's learning mechanism transitions from \textit{memorizing} isolated environments (high effective rank, block-structured eNTK) to \textit{adapting} across them, and finally to completely \textit{ignoring} the task-irrelevant nuisance features (collapsing states via PCA, rising SNR, and state CKA alignment) to maximize OOD generalization.}
    \label{fig:diagnostics}
    \end{figure}

To develop these diagnostic methods,
we will begin with a ``subproblem'' of a pick-and-place task:
training a CNN to output the size of a square in an image.
We will expand to robotic policies afterward,
but it will be clearer to start with this warm-up.
We will vary the number of colors considered for the square as $N_{\text{color}} \in \{1, 2, 3, 4, 7, 11, 17, 27\}$. For each value of $N_{\text{color}}$, we randomly sample 10 distinct color subsets to evaluate the performance variance under identical levels of data variation.

We construct 27 data probes representing different square colors, and compute the representation kernel matrix for each trained model configuration.
\Cref{fig:phase-transition} shows these matrices.
We can see that when the model is trained on only a few colors,
the kernel matrix often exhibits a clear block structure,
indicating that the model is treating these colors as isolated, distinct cases.
That is, training on a blue square may barely change the model's behavior on red squares.
We call this a \textit{memorize} regime,
as the model memorizes what to do for different situations,
rather than generalizing across them.
As the model is trained on more colors,
this block structure flattens
into an \emph{adapt} regime.

This shift is quantitatively captured in \cref{fig:erank}, where the \textbf{effective rank} \citep{erank} of the kernel matrices first increases as it is trained on more colors, then decreases to treat all colors similarly.

We compute both the in-distribution and out-of-distribution success rates across all $N_{\text{color}}$ values, averaging the results across the 10 random color subsets. As shown in \cref{fig:success-rate}, while nearly all randomization levels achieve near-perfect in-distribution success rates, they diverge significantly in their OOD performance.
Most notably, the surge in success at out-of-distribution prediction occurs around the same time as the kernel's effective rank flattens.

To investigate this phenomenon, we construct cross-wise paired data probes between the task-irrelevant square colors and the task-relevant states.
To better understand how the representation changes across the phase
transition, we first visualize the probe representations using a
two-dimensional PCA projection. Figure~\ref{fig:pca-snr} shows the resulting
projections for $N_{\text{color}}=4$ and $N_{\text{color}}=27$. When trained
on only a few colors, probes are broadly separated in the feature space,
including along the task-irrelevant color dimension. As the number of
training colors increases, probes with the same task-relevant state but
different colors become increasingly collocated, suggesting that the
representation becomes less sensitive to color while preserving
state-dependent structure.

For a more quantitative measurement, we build on distances in the feature space induced by the
representation kernel: $d(x,x')=\|\phi(x)-\phi(x')\|=\sqrt{K(x,x)+K(x',x')-2K(x,x')}$.
To attribute changes in these representations to individual sources of variation,
let $\mathcal{P}_a$ contain probe pairs that differ in factor $a$ while matching in factor $b$,
and let $\mathcal{P}_b$ contain the corresponding pairs that differ in $b$ while matching in $a$.
Letting
$\bar d_a=\operatorname{mean}_{(x,x')\in\mathcal{P}_a} d(x,x')$ and
$\bar d_b=\operatorname{mean}_{(x,x')\in\mathcal{P}_b} d(x,x')$,
we define the \emph{Factor Sensitivity Ratio} (FSR) as
$\mathrm{FSR}_{a:b}=\bar d_a/\bar d_b$.
FSR asks how strongly the representation changes when varying one factor
relative to another.
When one factor $s$ is task-relevant and another factor $n$ is
task-irrelevant, we call FSR by the familiar name of
\emph{Signal-to-Noise Ratio}:
$\mathrm{SNR}=\mathrm{FSR}_{s:n}=\bar d_s/\bar d_n$.
A larger SNR
indicates that task-relevant variation produces larger changes in the
representation relative to task-irrelevant variation.
In our color
experiment, $s$ is the square size, and $n$
the square color. As shown in Figure~\ref{fig:snr-across-n}, SNR
increases with $N_{\text{color}}$, consistent with the representation
becoming increasingly sensitive to size relative to color.

FSR and SNR provide a relative comparison between factors, but do not
indicate which factor is responsible for a change in the ratio. To
characterize the organization associated with each factor separately, we
use Centered Kernel Alignment (CKA;
\citealp{cortes2012algorithms,kornblith2019similarity}). For a factor $a$,
we measure the alignment of a label kernel
$L_a(x,x')=\mathbf{1}[a(x)=a(x')]$
with $K$:
$\mathrm{CKA}(K,L_a)=\mathrm{HSIC}(K,L_a)/\sqrt{\mathrm{HSIC}(K,K)\,\mathrm{HSIC}(L_a,L_a)} \in [0, 1]$,
where
$\mathrm{HSIC}(K,L) \propto \operatorname{tr}(KHLH)$
and
$H=I_m-\frac{1}{m}\mathbf{1}\mathbf{1}^{\top}$.
High CKA indicates that the representation structure is more similar for examples with the same value of factor $a$.
The state and color CKA curves in Figure~\ref{fig:cka-across-n}
provide a factor-specific view of the same transition captured by the SNR
in Figure~\ref{fig:snr-across-n}.

\section{Experimental Analyses}
\begin{figure*}[t]
    \begin{subfigure}[t]{0.58\textwidth}
        \vspace{0pt}  
        \centering
        \begin{minipage}{\textwidth}
            \centering
            \caption{\textbf{Predictive Validation of NTK Geometry.} Spearman Rank Correlation ($\rho$) and P-values between NTK cosine similarity and empirical success rates across 27 color candidates.}
            \label{tab:ntk_correlation}
            \small
            \begin{tabular}{lcc}
                \toprule
                Training Cube & Spearman Rho ($\rho$) & p-Value \\ \midrule
                black          & 0.7065 & 0.0000 \\
                white          & 0.9165 & 0.0000 \\
                black\_white    & 0.8189 & 0.0000 \\
                all\_random      & 0.5712 & 0.0019 \\ \bottomrule
            \end{tabular}
        \end{minipage}

        \vspace{0.3cm}

        \begin{minipage}{\textwidth}
            \centering
            \caption{\textbf{Metric evaluation across SNR, State CKA and Color CKA}. Bold values indicate superior performance.}
            \label{tab:training_analysis}
            \footnotesize
            \setlength{\tabcolsep}{4pt}
            \begin{tabular}{llccc}
                \toprule
                \textbf{Background} & \textbf{Cube} & \textbf{SNR} $\uparrow$ & \textbf{State CKA} $\uparrow$ & \textbf{Color CKA} $\downarrow$ \\
                \midrule
                Pure & Pure & 3.0404   & \textbf{0.6620} & 0.0576  \\
                Colorful & Pure & \textbf{3.4941} & 0.6437 & \textbf{0.0280}  \\
                \midrule
                Pure & Pure & 2.5394  & \textbf{0.7164} & 0.0849  \\
                Pure & Colorful & \textbf{4.0933} & 0.6692 & \textbf{0.0167}  \\
                \bottomrule
            \end{tabular}
        \end{minipage}
    \end{subfigure}
    \hfill
    \begin{subfigure}[t]{0.38\textwidth}
        \vspace{0pt}  
        \centering
        \includegraphics[width=\textwidth]{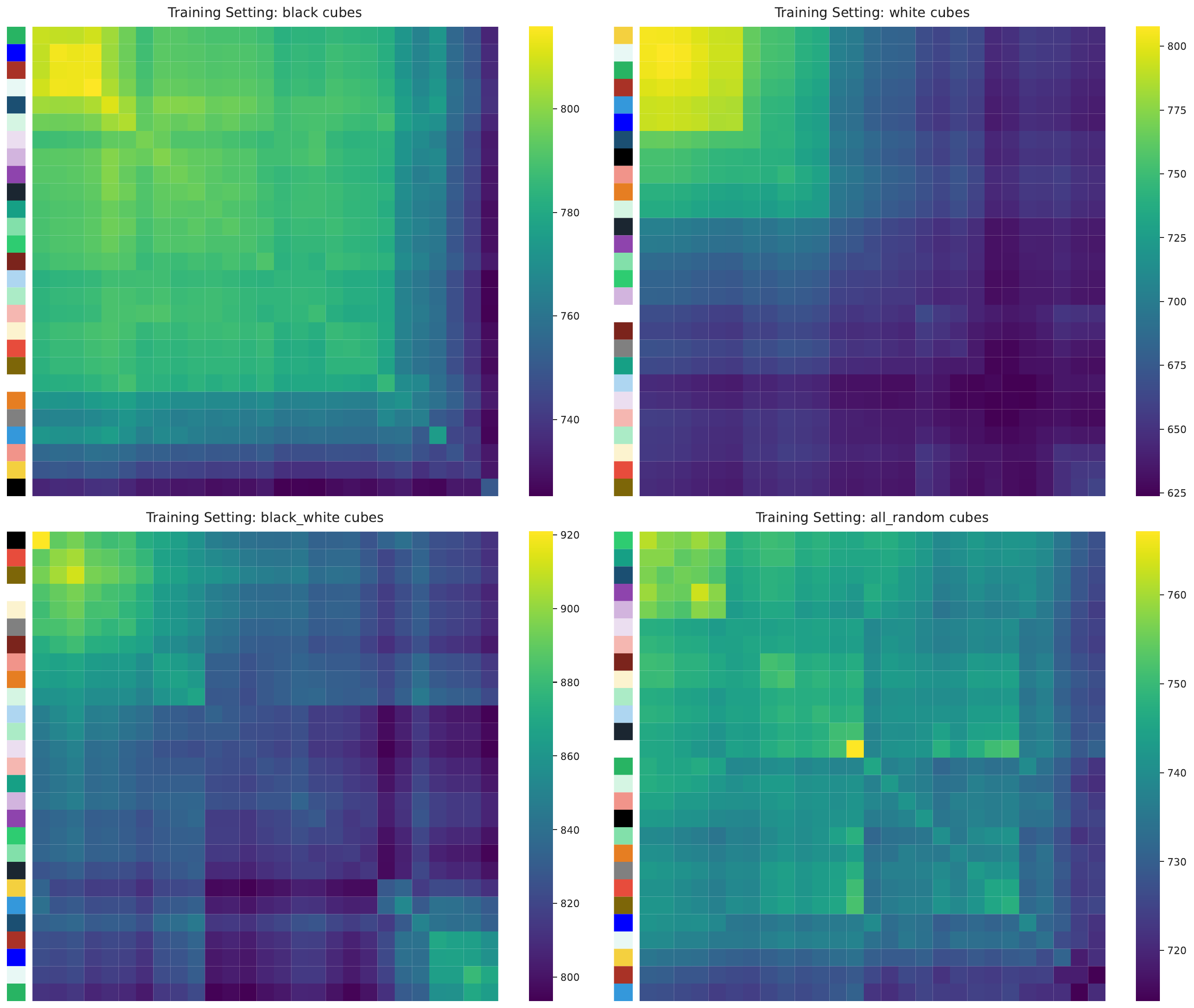}
        \caption{\textbf{NTK similarity for cube-based training settings.}
Hierarchical clustering reorders rows and columns for readability without
changing the kernel. Settings: black cube only (top left), white
cube only (top right), black and white cubes (bottom left), and random color cubes
(bottom right).}
        \label{fig:clustered_ntk_results}
    \end{subfigure}
    \caption{CNN policies for a ManiSkill pick-cube task.}
    \label{fig:overall}
\end{figure*}

\paragraph{RL Policies}
We now turn to
a pick-cube task in the ManiSkill simulator \citep{Su-maniskill-simulation-rss-2025}, analyzing CNN policies trained via reinforcement learning across domain randomization intensities. 

To investigate the internal learning mechanisms of the policy, we first establish the predictive power of the kernel geometry as a valid diagnostic proxy for empirical performance. 
\Cref{tab:ntk_correlation} evaluates the Spearman rank correlation between the kernel cosine similarity and empirical success rates across unseen evaluation colors.
The high correlation across four training DR settings
justifies using kernel statistics to reliably diagnose policy learning without requiring exhaustive evaluation rollouts.

We investigate the impact of visual diversity on representation structure by
analyzing models trained under different cube-color configurations.
Visualizing the resulting kernel matrices as heatmaps (Figure~\ref{fig:clustered_ntk_results})
reveals a distinct structural phase transition from a \emph{memorize} to an
\emph{adapt} regime. Insufficient DR yields prominent block-diagonal
structures, demonstrating a memorization regime where the model treats
different visual distributions as isolated, independent sub-tasks. Conversely,
sufficient DR (e.g., the \texttt{random\_color} setting) smooths out the kernel
structure, indicating an \emph{adapt}-style regime.

We also investigate whether policy learning suppresses task-irrelevant features by evaluating training performance under changing cube and background colors. For this analysis, we construct data probes cross-wise paired between scene states and colors. Quantitative alignment metrics, detailed in Table~\ref{tab:training_analysis}, show that sufficient background and object DR lead to an \textit{ignore} regime, which collapses irrelevant visual variance and lead to out-of-distribution (OOD) policy robustness.

\paragraph{Transformer Visuomotor Policies} 
We train ACT policies for peg insertion under increasing table-texture
variation. For tractability, we fix the CVAE latent $z$ to its prior mean, as at inference, and compute the eNTK with respect to a LoRA adapter's parameters. Validation loss can be misleading for OOD robustness:
a policy trained from scratch can overfit the in-distribution textures while
achieving low validation loss. Indeed, the policy trained without variation
attains the {lowest} validation loss, but the \emph{worst} OOD success
(Table~\ref{tab:act_single}). In contrast, both SNR diagnostics correctly
track OOD success, from $0.05$ to $0.43$.

We further vary two factors, texture and color, either individually or jointly,
and evaluate on corresponding single-factor and joint shifts
(Table~\ref{tab:act_joint}). Single-factor variation mainly improves
generalization to that factor, whereas joint variation is needed for joint
shifts, consistent with prior work \citep{gao2024}.
SNR on joint probes correctly orders all
four training conditions by joint OOD success, showing that the diagnostic also
captures the effect of training-data composition.

\begin{table}[t]
\centering
\footnotesize
\setlength{\tabcolsep}{4pt}
\renewcommand{\arraystretch}{1.1}
\begin{subtable}[t]{0.40\textwidth}
\centering
\begin{tabular}{l c cc c}
\toprule
 & & \multicolumn{2}{c}{SNR$\uparrow$} & \\
\cmidrule(lr){3-4}
DR & Val.\ loss$\downarrow$ & act. & eNTK & OOD$\uparrow$ \\
\midrule
none & \textbf{0.098} & 0.37 & 0.54 & 0.05 \\
mid  & 0.106 & 0.59 & 0.74 & 0.29 \\
high & 0.121 & \textbf{0.79} & \textbf{0.89} & \textbf{0.43} \\
\bottomrule
\end{tabular}
\caption{Single factor (table texture).}
\label{tab:act_single}
\end{subtable}%
\hfill
\begin{subtable}[t]{0.56\textwidth}
\centering
\begin{tabular}{l ccc ccc}
\toprule
 & \multicolumn{3}{c}{eNTK SNR (probes)$\uparrow$} & \multicolumn{3}{c}{OOD success$\uparrow$} \\
\cmidrule(lr){2-4}\cmidrule(lr){5-7}
DR & tex. & col. & joint & tex. & col. & joint \\
\midrule
none    & 0.45 & 1.19 & 0.42 & 0.05 & 0.35 & 0.00 \\
color   & 0.45 & \textbf{1.60} & 0.51 & 0.08 & \textbf{0.81} & 0.04 \\
texture & \textbf{0.62} & 1.39 & 0.60 & \textbf{0.43} & 0.39 & 0.24 \\
joint   & 0.56 & \textbf{1.60} & \textbf{0.65} & 0.31 & 0.49 & \textbf{0.35} \\
\bottomrule
\end{tabular}
\caption{Two factors (texture, color), varied separately or jointly.}
\label{tab:act_joint}
\end{subtable}
\caption{\textbf{eNTK SNR tracks OOD success for ACT policies (peg insertion).}
(a) The no-DR policy has the lowest validation loss but the worst OOD success; both SNRs increase with DR.
(b) Joint-probe SNR orders all four training conditions by joint OOD success.}
\label{tab:act}
\end{table}

 \paragraph{Scaling to Vision-Language-Action Policies}
We next apply our diagnostics to large Vision-Language-Action (VLA) policies on the LIBERO
and RoboTwin benchmarks, computing the eNTK over parameter-efficient LoRA
adapters as in \cref{sec:kernels}. At this scale, the same FSR statistic supports two practical uses: ranking models by robustness and detecting shortcut learning, both without evaluation rollouts.

\begin{figure*}[t]
    \centering
    \begin{subfigure}[t]{0.28\textwidth}
        \centering
        \begin{subfigure}{0.46\textwidth}
            \includegraphics[width=\textwidth]{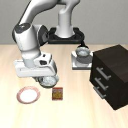}
        \end{subfigure}
        \hfill
        \begin{subfigure}{0.46\textwidth}
            \includegraphics[width=\textwidth]{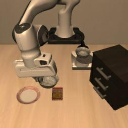}
        \end{subfigure}

        \vspace{0.5em}

        \begin{subfigure}{0.46\textwidth}
            \includegraphics[width=\textwidth]{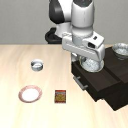}
        \end{subfigure}
        \hfill
        \begin{subfigure}{0.46\textwidth}
            \includegraphics[width=\textwidth]{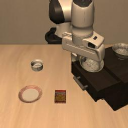}
        \end{subfigure}
        
        \caption{Lighting perturbations.}
        \label{fig:lighting_grid_sub}
    \end{subfigure}
    \qquad
    \begin{subfigure}[t]{0.60\textwidth}
    \centering
    \begin{tabular}{lcc}
    \toprule
    Model & eNTK SNR & OOD Success \\
    \midrule
    OpenVLA     & $2.802 \pm 0.176$ & \phantom{0}8.1\% \\
    RIPT-VLA    & $3.511 \pm 0.195$ & 88.4\% \\
    OpenVLA-OFT & $3.534 \pm 0.156$ & 88.7\% \\
    \bottomrule
    \end{tabular}
    \caption{
    SNR (mean $\pm$ std dev)
    indicates that
    \texttt{OpenVLA} is the most lighting-sensitive,
    agreeing with its empirical behavior on the LIBERO-plus benchmark \citep{fei25libero-plus}.}
    \label{fig:robust_robot_selection}
    \end{subfigure}
    \caption{OpenVLA models evaluated on LIBERO under lighting perturbations.}
\end{figure*}

As we saw that randomizing task-irrelevant factors increases SNR and improves OOD performance,
we use this insight we use this insight for zero-shot model selection. We treat lighting as
a task-irrelevant nuisance and task state as the task-relevant signal,
construct lighting-perturbation probes (Figure~\ref{fig:lighting_grid_sub}),
and compute $\mathrm{SNR} = \mathrm{FSR}_{\mathrm{state:lighting}}
= \bar d_{\mathrm{state}} / \bar d_{\mathrm{lighting}}$ for the original
OpenVLA \citep{Finn-openvla-corl-2024}, the fine-tuned OpenVLA-OFT
\citep{Liang-vla-finetuning-rss-2025} and RIPT-VLA~\cite{tan2025interactive}. A higher SNR means the representation
is more sensitive to state than to lighting. The SNR ordering of the models
matches their empirical OOD lighting robustness on the LIBERO-plus benchmark
\citep{fei25libero-plus}.

 \begin{figure}[t]
    \centering
    \begin{subfigure}[b]{0.33\textwidth}
        \centering
        \includegraphics[width=0.98\linewidth]{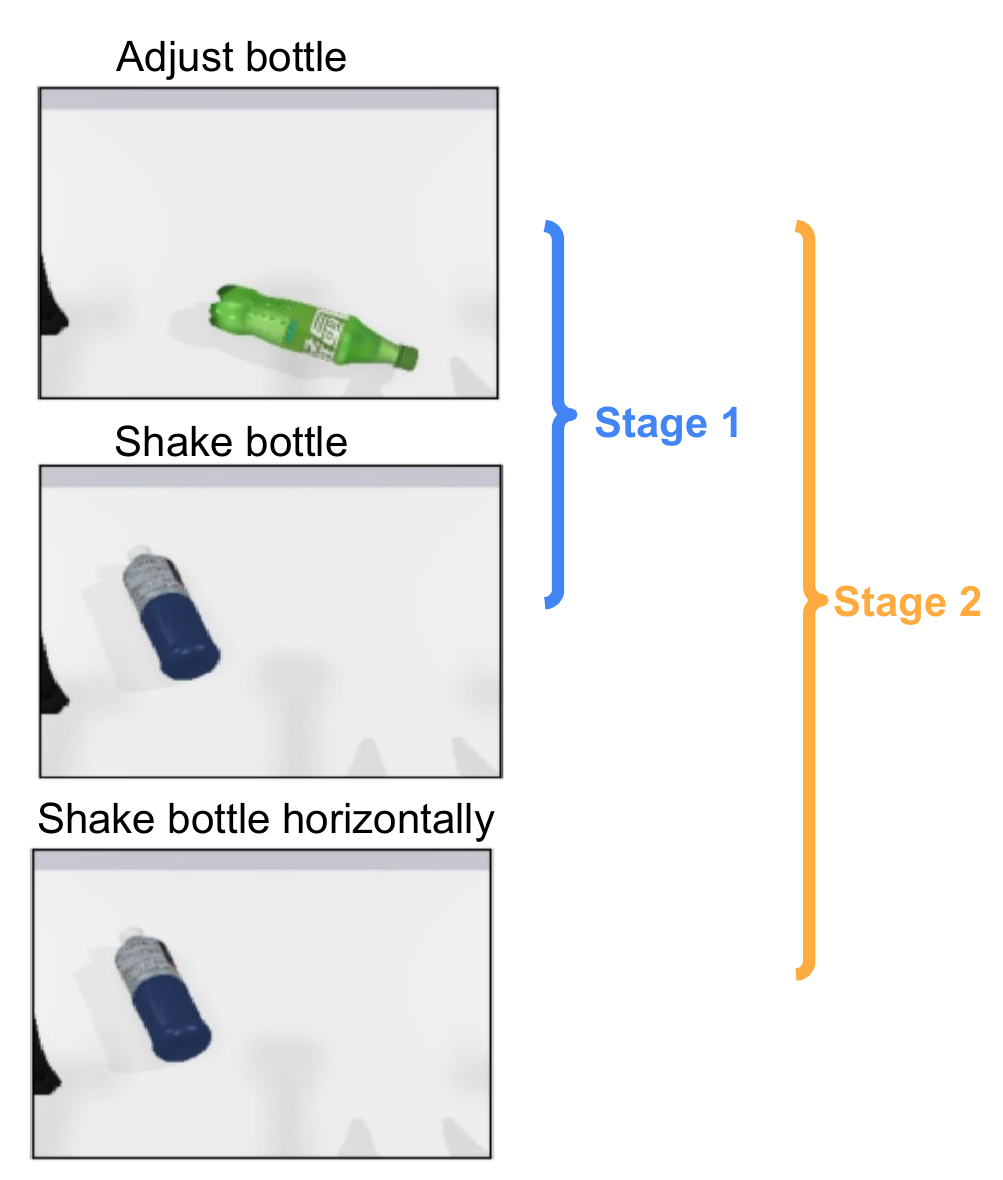}
        \caption{Training setting illustration.}
        \label{fig:shortcut_illustration}
    \end{subfigure}
    \hfill
    \begin{subfigure}[b]{0.65\textwidth}
        \centering
        \includegraphics[width=0.98\linewidth]{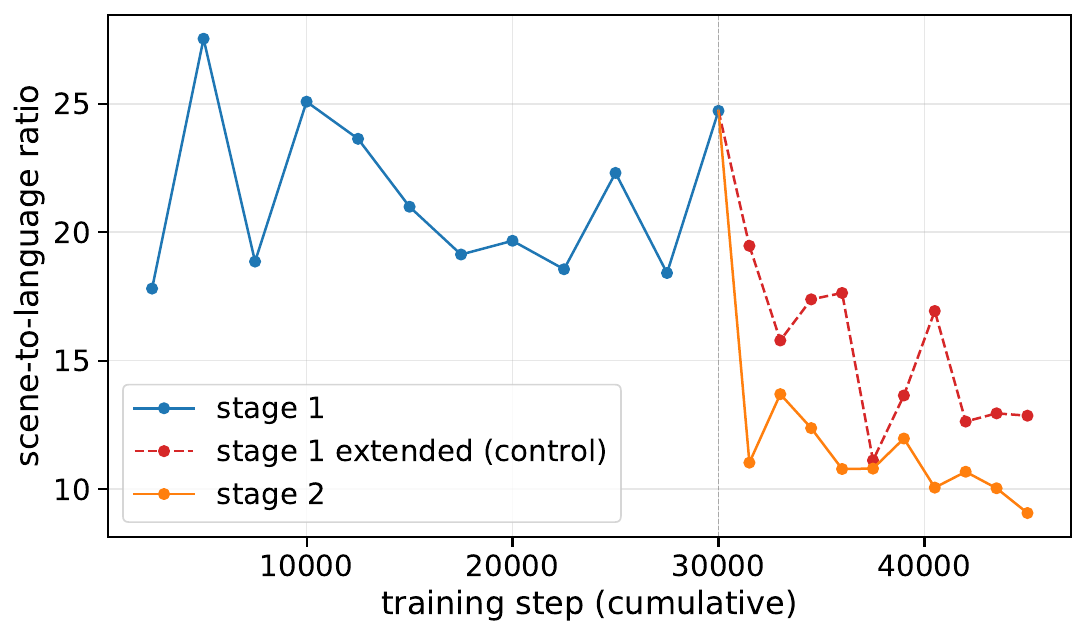}
        \caption{Scene to language ratio for two training stages.}
        \label{fig:shortcut_fsr}
    \end{subfigure}
    \caption{\textbf{Shortcut Learning Detection via Factor Sensitivity Ratio.}
(a) Training setting illustrating the visual--textual confound in Stage~1 and
its resolution in Stage~2. (b) Scene-to-language Factor Sensitivity Ratio
(FSR) across training, computed as the representation change induced by scene
variation relative to that induced by language variation. The decrease in FSR
after introducing the third task indicates reduced relative sensitivity to
scene variation and increased sensitivity to language variation.}
    \label{fig:shortcut_learning}
\end{figure}

Shortcut learning~\citep{xing2025shortcut} occurs when a policy relies on task-irrelevant cues rather than task-relevant information. To detect visual shortcut learning without task-level evaluation, we apply the FSR to compare representation sensitivity to scene and language variation. We evaluate OpenVLA-OFT on three RoboTwin tasks~\citep{chen2025robotwin20scalabledata}: adjust bottle, shake bottle, and shake bottle horizontally (Fig.~\ref{fig:shortcut_illustration}).

Stage 1 trains on the first two tasks, where each language instruction is perfectly confounded with a particular scene configuration. The model can therefore solve the training tasks by relying only on visual cues. We measure this using the scene-to-language ratio, $\mathrm{FSR}_{\mathrm{scene:language}}$.

Stage 2 introduces the third task, which breaks this visual--textual correlation and requires language to distinguish otherwise similar visual inputs. As shown in Fig.~\ref{fig:shortcut_fsr}, the FSR decreases after introducing the third task, indicating reduced relative sensitivity to scene variation and increased sensitivity to language. This drop is not simply a consequence of longer training. A control that continues first-phase training for the same 15k steps stays at 15.0, and the second phase is lower at all 10 step-matched checkpoints. FSR provides a representation-level signal of reduced visual shortcut reliance, without requiring task-level evaluation rollouts.

\paragraph{Independent Variation Beats Confounded Variation at Equal Data Budget}
\label{sec:independent_variation}

\begin{figure}[t]
  \centering
  \begin{subfigure}{\linewidth}
    \centering
    \includegraphics[width=\linewidth]{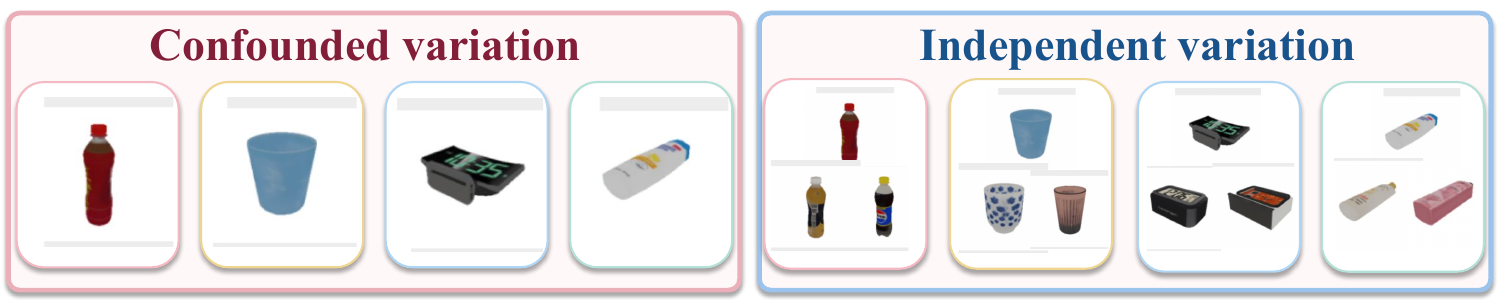}
    \caption{Data variation design. Confounded variation ties shape and texture,
             whereas independent variation varies texture across objects with
             similar shapes within a category.}
    \label{fig:var-design}
  \end{subfigure}

  \vspace{0.8em}
  \newsavebox{\ntkbox}
\sbox{\ntkbox}{\includegraphics[width=0.44\linewidth]{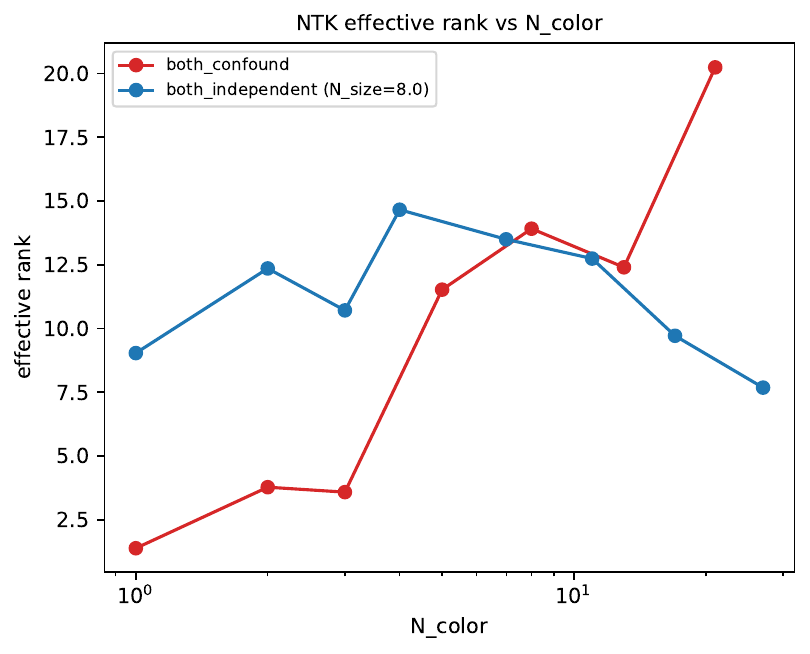}}

  \begin{subfigure}[t]{0.44\linewidth}
    \centering
    \adjustbox{valign=t}{\usebox{\ntkbox}}
    \caption{Effective rank during training on toy environment. Independent variation shows an initial rise followed by a decline, marking a transition toward an \emph{adapt}-style regime, while confounded variation keeps rising.}
    \label{fig:var-erank}
  \end{subfigure}\hfill
  \begin{subfigure}[t]{0.53\linewidth}
    \centering
    \begin{adjustbox}{minipage=[c][\ht\ntkbox][c]{\linewidth}, valign=t}
      \centering
      \small
      \begin{tabular}{lccc}
        \toprule
        Evaluation      & Confounded & Independent     & $\Delta$  \\
        \midrule
        In-distribution & 0.0605     & 0.0587          & $-0.0018$ \\
        Novel category  & 0.0846     & \textbf{0.0755} & $-0.0091$ \\
        \bottomrule
      \end{tabular}
    \end{adjustbox}
    \caption{Mean action error on evaluation categories. Both conditions use the
             same number of training episodes. Independent variation achieves lower
             error on novel categories (0.0755 vs.\ 0.0846), while in-distribution
             performance is similar (0.0587 vs.\ 0.0605).}
    \label{fig:var-error}
  \end{subfigure}

  \caption{\textbf{Independent variation improves generalization at equal data budget.}}
  \label{fig:variation}
\end{figure}

Confounded variation does not necessarily cause shortcut learning, but it can introduce performance penalties. In Fig.~\ref{fig:var-erank}, erank transition on toy environment (Sec.~\ref{sec:setup}) provides initial intuition: confounded variation remains more \emph{memorize}-like, whereas independent variation transitions toward \emph{adapt}.

We study this question in RoboTwin Place Object Scale task, where the policy places objects on a scale. We vary object shape, which is task-relevant, and texture, which is task-irrelevant. For each category, objects have similar but non-identical shapes. We compare \emph{confounded variation}, which fixes one object-texture pair per category, with \emph{independent variation}, which samples multiple objects with different textures within each category (Figure~\ref{fig:var-design}). Both settings use the
same data budget, i.e., the same number of training episodes and therefore
the same data-collection effort. On novel object categories, independent variation achieves lower action error, while the two conditions are indistinguishable in-distribution (Figure~\ref{fig:var-error}).

\paragraph{Representation Kernel Comparison}

We previously argued that the eNTK is a reasonable default representation kernel,
rather than universally superior to activation-based alternatives. We compare
it with the last-hidden-layer and output kernels on two diagnostics where
representation choice matters. Since metric magnitudes are not directly
comparable across representation spaces, we focus on qualitative trends within
each representation.

We repeat the shortcut-detection study on 80 grasp-frame probes. For this experiment, we use the
scene-to-language Factor Sensitivity Ratio (FSR),
$\bar d_{\mathrm{scene}}/\bar d_{\mathrm{lang}}$, to quantify the relative
sensitivity of each representation to the two factors. Both eNTK and
last-hidden representations detect the alleviated shortcut at all 10
matched checkpoints ($p=0.002$), whereas the output kernel does not
($4/10$, $p=0.16$; Table~\ref{tab:kernel_shortcut}). At the grasp
frame, both instructions require the same action, so the output representation
collapses the language conditions. Thus, output-space probes cannot reveal
reliance on a factor that does not yet affect the action.

We next use a bimanual task in which the two arms share all layers except their
final heads. The last-hidden representation is therefore identical for both
arms and cannot capture the arm-specific decoupling introduced by the heads.
Here, each arm has a natural task-relevant factor (its own-peg pose) and a
task-irrelevant factor (the partner-peg pose), so we use the
signal-to-noise ratio (SNR), which is the corresponding special case of FSR.
For each arm, the signal distance is placed in the numerator, with the ratio
inverted when necessary so that $\mathrm{SNR}\geq 1$. The last-hidden SNR
therefore remains near 1 because the two arms have identical last-hidden
representations, while the output kernel detects the head-specific decoupling
directly, with SNR increasing from about 2 to 48 for both arms. The eNTK
captures the same trend more weakly (Table~\ref{tab:kernel_bimanual}).

The three representations have complementary blind spots: output space can
miss distinctions that do not yet affect the action, while the last hidden
layer can miss distinctions introduced in the head. The eNTK aggregates
information across layers and has a direct physical interpretation through
parameter-update effects, making it a reasonable default diagnostic
representation in this paper. In practice, however, activation and output
representations are substantially cheaper to compute and can be used within
the same diagnostic framework. The representation can therefore be chosen
case by case according to which aspects of the data it preserves.

\label{sec:representation-comparison}
\begin{table}[t]
\centering
\footnotesize
\setlength{\tabcolsep}{3.5pt}
\begin{subtable}[t]{0.43\textwidth}
\centering
\begin{tabular}{lcccc}
\toprule
space & Stage 1 & Stage 2 & control & S2 $<$ ctrl \\
\midrule
eNTK        & 21.4 & 11.0 & 15.0 & 10/10 \\
last hidden & 8.5  & 7.0  & 9.5  & 10/10 \\
output      & 148  & 136  & 127  & 4/10  \\
\bottomrule
\end{tabular}
\caption{Scene-to-language FSR $\bar d_{\text{scene}}/\bar d_{\text{lang}}$
on 80 grasp-frame probes (lower = more reliance on language).}
\label{tab:kernel_shortcut}
\end{subtable}\hfill
\begin{subtable}[t]{0.54\textwidth}
\centering
\begin{tabular}{llcccc}
\toprule
& & \multicolumn{3}{c}{training epoch} & \\
\cmidrule(lr){3-5}
arm & space & 0 & 100 & 1900 & trend \\
\midrule
\multirow{3}{*}{A vs B}
  & output      & 1.9  & 15.0 & 47.8 & $\uparrow$ \\
  & last hidden & 1.05 & 0.99 & 1.02 & $\Downarrow$ \\
  & eNTK (all)  & 1.02 & 1.07 & 1.14 & $\uparrow$ \\
\midrule
\multirow{3}{*}{B vs A}
  & output      & 1.4  & 15.8 & 47.7 & $\uparrow$ \\
  & last hidden & 0.95 & 1.01 & 0.98 & $\Uparrow$ \\
  & eNTK (all)  & 0.97 & 1.08 & 1.12 & $\uparrow$ \\
\bottomrule
\end{tabular}
\caption{Own-peg vs.\ partner-peg SNR in bimanual peg picking
(CNN-MLP, $10\times10$ pose grid, 3 seeds).}
\label{tab:kernel_bimanual}
\end{subtable}
\caption{\textbf{Representation kernels have complementary blind spots.}
(a) The output kernel misses the shortcut because both instructions
require the same grasp-frame action.
(b) The arms share all layers except their final heads, so the
last-hidden kernel cannot see arm-specific structure.}
\label{tab:kernel}
\end{table}

\paragraph{Hardware Validation}
\label{sec:hardware}

\begin{figure}[t]
  \centering
  \begin{subfigure}{0.48\linewidth}
    \includegraphics[width=\linewidth]{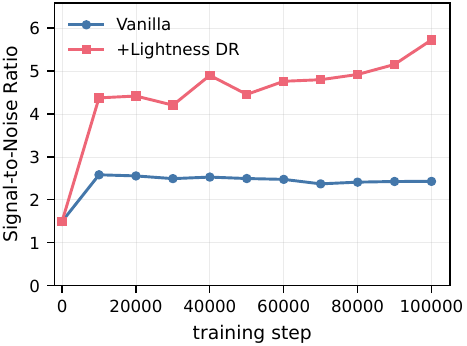}
    \caption{Signal-to-Noise ratio over training.}
    \label{fig:snr-curve}
  \end{subfigure}
  \hfill
  \begin{subfigure}{0.48\linewidth}
    \centering
    \includegraphics[width=0.48\linewidth]{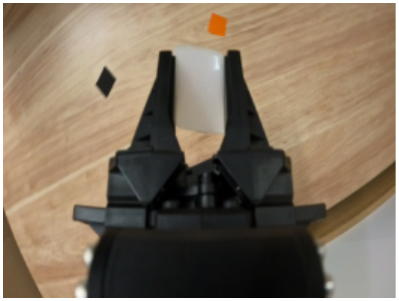}\hfill
    \includegraphics[width=0.48\linewidth]{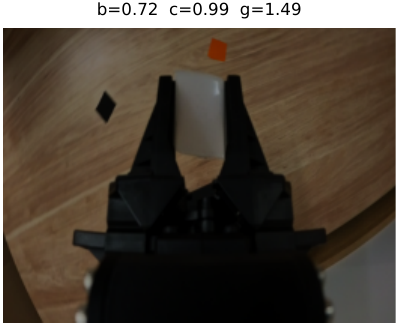}\\[2pt]
    \includegraphics[width=0.48\linewidth]{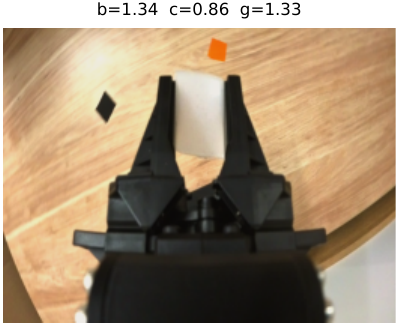}\hfill
    \includegraphics[width=0.48\linewidth]{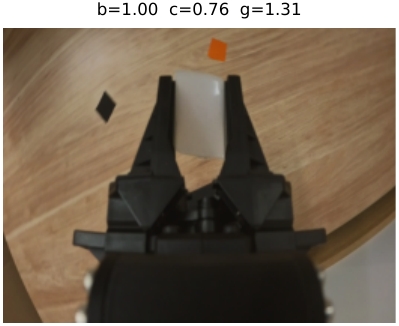}
    \caption{Lightness DR samples.}
    \label{fig:lightness-samples}
  \end{subfigure}
  \caption{Lightness domain randomization increases the SNR from roughly $2.5$ to $5.5$, indicating features become more separable across the augmented inputs. Samples are annotated with brightness $b\!\in\![0.6,1.4]$, contrast $c\!\in\![0.7,1.3]$, and gamma $g\!\in\![0.7,1.5]$, applied per-sample in that order.}
  \label{fig:lightness-snr}
\end{figure}

We validate the factor-conditioned kernel diagnostics on a hardware
cube pick-and-place task using an ACT policy~\citep{act}. Lightness DR increases the training SNR from 2.5 to 5.5, indicating greater relative sensitivity to task-relevant state than to lighting variation
(Figure~\ref{fig:snr-curve}). This improvement is reflected in
real-world robustness: success increases from $1/10$ to $8/10$ under
bright lighting and from $1/10$ to $6/10$ under dark lighting, while
in-distribution success increases from $6/10$ to $10/10$. Thus, the
diagnostic tracks the improvement in visual robustness on hardware;
full evaluation details are provided in Appendix~\ref{sec:hardware-app}.

\section{Limitations}
\label{sec:limitations}

\paragraph{Limited statistical power.}
``Memorize,'' ``adapt,'' and ``ignore'' are operational descriptions of kernel structure, not proven causal mechanisms, and the statistics depend on the probe design, factor parameterization, checkpoint, and representation. We therefore view our results as supporting evidence rather than conclusive validation. The RL, RoboTwin, and hardware policies are single runs, and the hardware study uses only 10 trials per condition, so it is a consistency check on one comparison between two policies, not statistical validation. The effective-rank transition is also shown only on the toy environment (Figure~\ref{fig:var-erank}) and not on RoboTwin (Figure~\ref{fig:var-error}), since it requires a series of checkpoints trained with different amounts of variation, which is computationally heavy.

\paragraph{Preliminary probe construction.}
The framework requires probe sets defined around known factors and cannot discover unknown sources of variation. Probes drawn from the evaluation distribution are ideal. Probes from outside both the training and evaluation distributions may also reveal OOD generalization ability, but they are not guaranteed to reflect performance on a specific OOD evaluation. Our probe construction is preliminary: we provide neither a formal specification for parameterizing or choosing probes nor a quantitative guarantee of statistical confidence in the resulting metrics. A structured treatment of probe design, with quantitative confidence estimates, is a natural next step.

\paragraph{Kernel choice and comparability.}
Kernel metric magnitudes are not directly comparable across representation spaces, so our conclusions rely on qualitative trends and within-kernel comparisons. We compare activation, output, and eNTK kernels and show that each has its own advantages and blind spots, but we do not provide strict, off-the-shelf guidance on which to use when.

Our goal is to introduce the ``memorize, adapt, ignore'' framework as an initial step and to draw attention to an underexplored connection between representation-kernel metrics and the learning regimes of robot policies. Establishing how reliably the metrics generalize across all tasks, nuisance factors, and probe distributions would require a more extensive study than this paper provides. Still, the experiments span a wide range of settings (a toy illustration, RL with CNN policies, imitation learning with transformer visuomotor policies, VLA model selection and shortcut detection, and hardware), and across them the kernel metrics consistently distinguish different learning behaviors. We believe this breadth supports our core claim that kernel metrics are a useful diagnostic for robot learning, even though each individual setting has limited coverage.

\section{Conclusion}
\label{sec:conclusion}
We present a representation-kernel framework for diagnosing how robot policies respond to training-data variation. Across RL, imitation learning, and VLA models, our diagnostics distinguish three learning regimes: memorizing isolated conditions, adapting across task-relevant variation, and ignoring task-irrelevant variation. Kernel-based SNR and factor sensitivity provide signals for assessing OOD robustness and shortcut learning without relying solely on task-level evaluation.

Our results show that training-data design systematically shapes these learning regimes. Increasing relevant variation can drive policies from memorization toward adaptation, while variation in task-irrelevant factors promotes invariance. Independently varying factors also improves generalization over confounded variation at the same data budget. These patterns extend across model classes and are reflected in physical hardware performance, supporting representation kernels as practical diagnostics for data design and model analysis.

More broadly, our contribution is a robotics-focused framework that brings existing tools together to address practical challenges in robot learning. Limited data, costly rollouts, and interpretable axes of generalization make representation-level probes a practical way to study how training-data variation shapes internal learning mechanisms, complementing task-level evaluation. This perspective highlights the value of studying how robot policies learn internally, not only what they achieve, and opens opportunities to use representation-level diagnostics to inform data collection, model selection, and robustness evaluation.

\bibliographystyle{plainnat}
\bibliography{references}

\appendix
\section{Appendix}
\subsection{Hardware Evaluation Details}
\label{sec:hardware-app}

\begin{figure}[!htbp]
  \centering
  \begin{subfigure}{0.31\linewidth}
    \includegraphics[width=\linewidth]{img/hardware/real_base.pdf}
    \caption{Base}
    \label{fig:real-base}
  \end{subfigure}\hfill
  \begin{subfigure}{0.31\linewidth}
    \includegraphics[width=\linewidth]{img/hardware/real_bright.pdf}
    \caption{Bright (OOD)}
    \label{fig:real-bright}
  \end{subfigure}\hfill
  \begin{subfigure}{0.31\linewidth}
    \includegraphics[width=\linewidth]{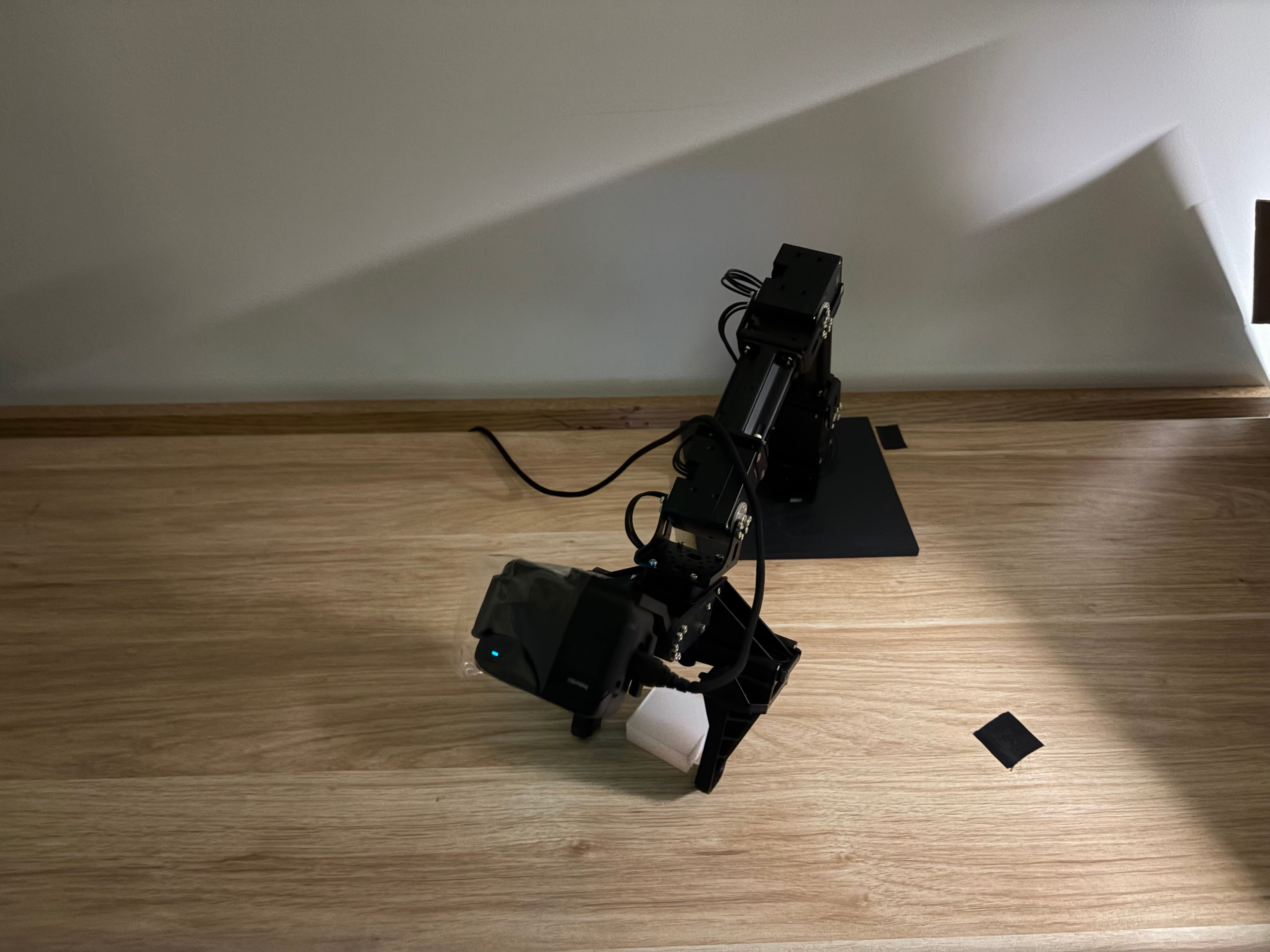}
    \caption{Dark (OOD)}
    \label{fig:real-dark}
  \end{subfigure}
  \caption{Real-world evaluation conditions. The Base setting matches
  training; Bright and Dark are out-of-distribution lighting
  variations.}
  \label{fig:real-eval-setup}
\end{figure}

We evaluate an ACT imitation learning policy~\citep{act} on a hardware
cube pick-and-place task under three lighting conditions
(Fig.~\ref{fig:real-eval-setup}): \emph{Base}, which matches the
training distribution, and \emph{Bright} and \emph{Dark}, which
introduce out-of-distribution (OOD) illumination changes. We compare a
baseline trained without Lightness DR against a policy trained with
Lightness DR. Each condition is evaluated over 10 trials.

Table~\ref{tab:real-eval} reports success rates. Without Lightness DR,
performance collapses from 6/10 to 1/10 under both OOD conditions. With
Lightness DR, the policy reaches 10/10 in-distribution and retains
8/10 and 6/10 under Bright and Dark lighting, respectively.

\begin{table}[!htbp]
  \centering
  \caption{Real-world success rates over 10 trials per condition.
  Parenthesized values give the relative drop from the Base condition.
  Lightness DR improves OOD success from 1/10 to 8/10 (Bright) and
  6/10 (Dark), while also improving in-distribution success.}
  \label{tab:real-eval}
  \begin{tabular}{lccc}
    \toprule
    & Base & Bright (OOD) & Dark (OOD) \\
    \midrule
    Without Lightness DR & 6/10 & 1/10 \scriptsize{($-$83\%)} & 1/10 \scriptsize{($-$83\%)} \\
    With Lightness DR    & \textbf{10/10} & \textbf{8/10} \scriptsize{($-$20\%)} & \textbf{6/10} \scriptsize{($-$40\%)} \\
    \bottomrule
  \end{tabular}
\end{table}


\subsection{eNTK vs.\ Activation Similarity as a Predictor of Transfer}
\label{sec:ntk-vs-act}

A natural alternative to the empirical neural tangent kernel (eNTK) is
to measure similarity between inputs directly in a network's
activation space. Here we test which of the two better predicts how
training on one input changes the model's output on another.

\paragraph{Setup.}
We use the square-size CNN of Sec.~\ref{sec:setup}, taking one
checkpoint per $N_{\mathrm{color}} \in \{1,\dots,27\}$. For each
checkpoint, we take a single small SGD step on a probe input $x$ and
measure the induced change $\Delta f(x')$ in the output on held-out
probes $x'$, yielding approximately 1.7k $(x, x')$ pairs per
checkpoint. We then compute the Spearman rank correlation between
$\Delta f(x')$ and each candidate similarity measure.

\paragraph{Similarity measures.}
The eNTK is
\begin{equation}
  K(x, x') = \nabla_\theta f(x)^{\top} \nabla_\theta f(x'),
\end{equation}
which, to first order in the step size $\eta$, governs the output
change on $x'$ after a gradient step on $x$:
\begin{equation}
  \Delta f(x') \approx -\eta\, K(x', x)\, \nabla_{f}\mathcal{L}\big(f(x)\big).
\end{equation}
For activation similarity, we compute the cosine similarity between
the representations of $x$ and $x'$ at every layer of the network,
reporting both the MLP head and the layer that achieves the highest
correlation (chosen post hoc, which favors this baseline).

\paragraph{Results.}
Table~\ref{tab:ki2_training_prediction} shows that the eNTK nearly
determines the measured transfer ($\rho = 0.97 \pm 0.06$). Activation
similarity is weak at every layer: even the best layer, selected with
access to the outcome, reaches only $\rho = 0.22 \pm 0.28$, and the MLP
head is close to uncorrelated with high variance across checkpoints
($\rho = 0.09 \pm 0.52$). Two inputs can therefore have similar
representations while being coupled only weakly through shared
parameters, and vice versa. Because transfer under gradient descent
depends on how parameter gradients align rather than on how
representations align, the eNTK is the appropriate object for
predicting cross-input effects of training.

\begin{table}[!htbp]
  \centering
  \caption{Predicting training transfer from similarity. Spearman $\rho$
  between similarity $\mathrm{sim}(x,x')$ and measured output change
  $\Delta f(x')$ after one SGD step on $x$, reported as mean $\pm$ s.d.\
  across 27 checkpoints ($\approx$1.7k pairs each). The best activation
  layer is chosen post hoc.}
  \label{tab:ki2_training_prediction}
  \begin{tabular}{lc}
    \toprule
    Similarity space & Spearman $\rho\,\big(\mathrm{sim}(x,x'),\, \Delta f(x')\big)$ \\
    \midrule
    \textbf{eNTK} $K(x,x')$  & $\mathbf{+0.97 \pm 0.06}$ \\
    \midrule
    Activations (best layer) & $+0.22 \pm 0.28$ \\
    Activations (MLP head)   & $+0.09 \pm 0.52$ \\
    \bottomrule
  \end{tabular}
\end{table}

\subsection{Separate Scene and Language CKA in the Shortcut Analysis}
\label{app:scene_lang}

\begin{figure}[t]
    \centering
    \includegraphics[width=\linewidth]{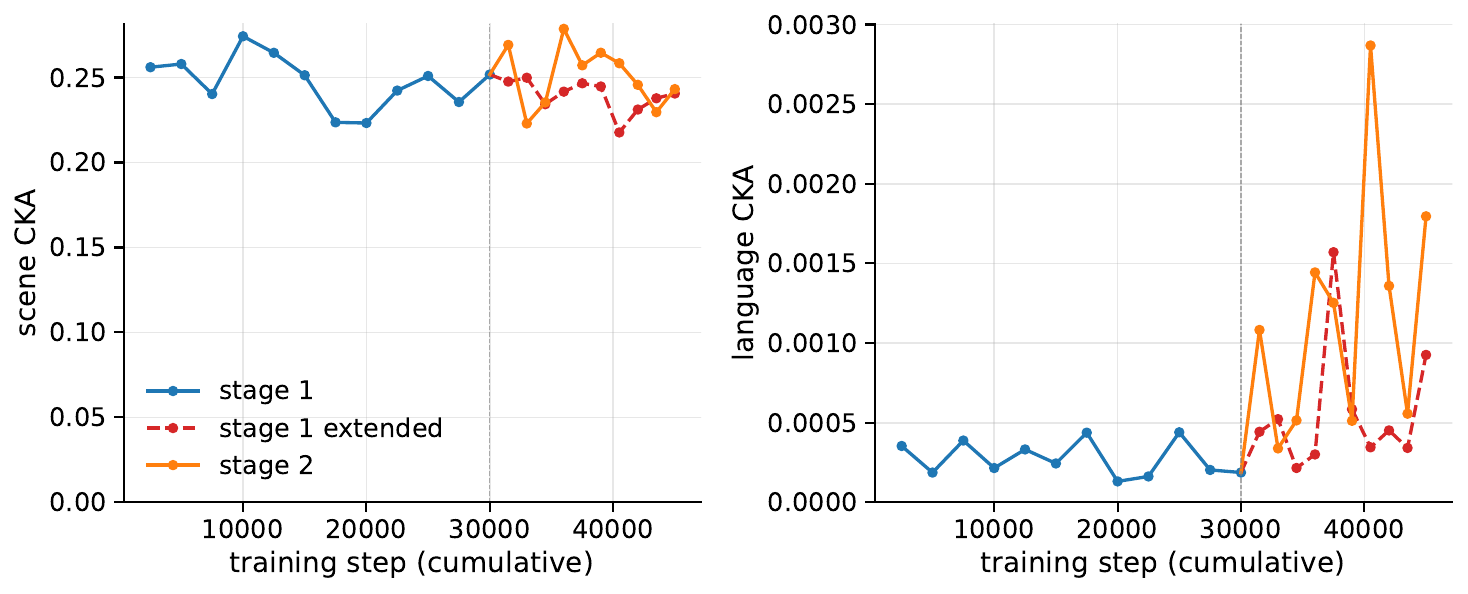}
    \caption{\textbf{Factor-specific decomposition of the Figure~\ref{fig:shortcut_fsr} shortcut analysis.}
    Linear CKA between the eNTK and the scene-label (left) and language-label (right)
    kernels on a fixed probe set from the two original tasks
    (40 scenes $\times$ \{adjust, shake\}, 80 probes), computed on the same raw kernel as
    the FSR curve in Figure~\ref{fig:shortcut_fsr}. Relative to the stage-1 mean, scene CKA stays within
    $0.9$--$1.1\times$ in all three runs, whereas language CKA rises to $4.3\times$ in
    stage~2 and to $2.1\times$ in the extended stage-1 control. The FSR decrease in
    Figure~\ref{fig:shortcut_fsr} is therefore driven by the language factor, not by a change in scene
    sensitivity, and the rise is larger when the third task is added than under
    continued training alone.}
    \label{fig:bottle_cka}
  \end{figure}

The scene-to-language FSR in Figure~\ref{fig:shortcut_fsr} is a ratio, so it cannot by itself show which factor changed.
To attribute the Stage~2 decrease to a specific factor, we compute factor-specific Centered Kernel Alignment (CKA; Section~3) between the eNTK and two label kernels: a scene-label kernel and a language-label kernel.
Both are computed with linear CKA on a fixed probe set drawn from the two original tasks (40 scenes $\times$ \{adjust, shake\}, 80 probes), using the same raw kernel as the FSR curve in Figure~\ref{fig:shortcut_fsr}.
Figure~\ref{fig:bottle_cka} reports both curves for Stage~1, Stage~2, and the step-matched Stage~1 control.

Relative to the Stage~1 mean, scene CKA stays within $0.9$--$1.1\times$ in all three runs, so the representation's alignment with the scene does not change after the third task is introduced.
In contrast, language CKA rises to $4.3\times$ its Stage~1 mean in Stage~2 and to $2.1\times$ in the extended Stage~1 control.
The decrease in scene-to-language FSR is therefore driven by the language factor rather than by a change in scene sensitivity.
The rise in language CKA is larger when the third task is added than under continued training alone, which is consistent with the third task, and not longer training, making the representation sensitive to language.
We note that the absolute language CKA values are small and the curve is noisy across checkpoints, so we interpret the ratio to the Stage~1 level rather than individual values.


\subsection{Stability of the VLA Model Ranking Across Probe Groups}
\label{app:probe_groups}

Table~\ref{tab:fig3b_probe_groups} lists the per-group SNR values behind the mean $\pm$ std in Figure~\ref{fig:robust_robot_selection}.

\begin{table}[t]
  \centering
  \caption{eNTK lighting SNR per probe group (Figure~\ref{fig:robust_robot_selection} resampling). Each group is 5 scenes $\times$ 5 lighting levels drawn from disjoint LIBERO-Spatial episodes; model, LoRA draw, and projection are held fixed. OpenVLA ranks
  lowest in every group (exact rank-sum $p = 1/3003$), while RIPT-VLA and OpenVLA-OFT swap order across groups (gap $-0.023 \pm 0.053$), consistent with their tie on LIBERO-plus.}
  \label{tab:fig3b_probe_groups}
  \begin{tabular}{lccc}
  \toprule
  Probe group & OpenVLA & RIPT-VLA & OpenVLA-OFT \\
  \midrule
  1 & \textbf{2.966} & 3.788 & 3.791 \\
  2 & \textbf{2.584} & 3.614 & 3.600 \\
  3 & \textbf{2.738} & 3.495 & 3.494 \\
  4 & \textbf{3.048} & 3.464 & 3.460 \\
  5 & \textbf{2.675} & 3.195 & 3.323 \\
  \midrule
  Mean $\pm$ std & $2.802 \pm 0.176$ & $3.511 \pm 0.195$ & $3.534 \pm 0.156$ \\
  Ranked lowest & 5/5 & 0/5 & 0/5 \\
  LIBERO-plus Light (\%) & 8.1 & 88.4 & 88.7 \\
  \bottomrule
  \end{tabular}
\end{table}

\end{document}